\documentclass[11pt,a4paper]{article}

\usepackage[acceptedWithA]{tacl2021v1}
\usepackage{microtype}

\usepackage{xspace,mfirstuc,tabulary}

\usepackage{times,latexsym}
\usepackage{url}
\usepackage{inconsolata}
\usepackage[T1]{fontenc}

\usepackage{latexsym}

\usepackage{comment}
\usepackage[disable]{todonotes}
\newcommand{\note}[4][]{\todo[author=#2,color=#3,size=\scriptsize,fancyline,caption={},#1]{#4}} 
\newcommand{\josef}[2][]{\note[#1]{josef}{blue!40}{#2}}

\usepackage{cleveref}
\usepackage{graphicx}
\usepackage{booktabs}
\crefname{section}{\S}{\S\S}
\crefname{table}{Tab.}{}
\crefname{figure}{Fig.}{}
\crefname{algorithm}{Alg.}{}
\crefname{equation}{Eq.}{Eq.}
\crefname{appendix}{App.}{}
\crefname{theorem}{Theorem}{}
\crefname{prop}{Proposition}{}
\crefname{cor}{Corollary}{}
\crefname{observation}{Observation}{}
\crefname{assumption}{Assumption}{}
\crefname{hypothesis}{Hyp.}{Hypotheses}
\crefformat{section}{\S#2#1#3}
\crefname{recommendation}{Recommendation}{}
\newtheorem{recommendation}{Recommendation}

\usepackage{emoji}
\newcommand{\ucambridge}{\emoji[twitter_emoji]{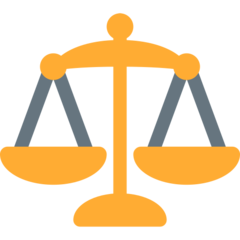}}
\newcommand{\eth}{\emoji[twitter_emoji]{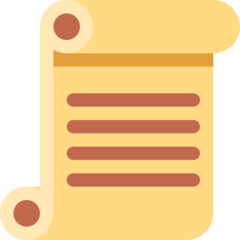}}

\title{The Ethics of Automating Legal Actors}

\author{Josef Valvoda$^{\ucambridge}$ Alec Thompson$^{\ucambridge}$
\textbf{Ryan Cotterell $^{\eth}$ Simone Teufel $^{\ucambridge}$} \\
$^{\ucambridge}$University of Cambridge
~\;~$^{\eth}$ETH Z\"{u}rich \\
  \{\texttt{\href{mailto:jv406@cam.ac.uk}{jv406}}, \texttt{\href{mailto:at808@cam.ac.uk}{at808}}, \texttt{\href{mailto:sht25@cam.ac.uk}{sht25}\}@cam.ac.uk} ~\;~\texttt{\href{mailto:ryan.cotterell@inf.ethz.ch}{ryan.cotterell@inf.ethz.ch}}\\ 
}

\date{}

\begin{document}

\maketitle

\begin{abstract}
The introduction of large public legal datasets has brought about a renaissance in legal NLP. 
Many of these datasets are comprised of legal judgements -- the product of judges deciding cases. 
This fact, together with the way machine learning works, means that several legal NLP models are models of judges.
While some have argued for the automation of judges, in this position piece, we argue that automating the role of the judge raises difficult ethical challenges, in particular for common law legal systems.
Our argument follows from the social role of the judge in actively shaping the law, rather than merely applying it.
Since current NLP models come nowhere close to having the facilities necessary for this task, they should not be used to automate judges.
Furthermore, even in the case the models could achieve human-level capabilities, there would still be remaining ethical concerns inherent in the automation of the legal process.
\end{abstract}

\section{Introduction}

This paper discusses the ethical aspects of using natural language processing (NLP) research to augment or replace the work of legal experts. 
It considers common law legal systems and demonstrates that many proposals for automation in the legal domain concentrate on the judiciary. 
Reflecting on this sentiment, we find that a number of popular legal datasets are focused on the text produced by judges.
Whilst we agree that there are potential benefits to the practical application of NLP to the legal domain, these applications face several ethical challenges. 
Some of these challenges are resolvable with technical advances, others, however, appear to be intrinsic to using any kind of automation.
We consider the two main legal actors, the judge and the lawyer and find that while automation of either can be beneficial, lawyer automation presents fewer challenges.\looseness-1

We begin by giving a brief legal background in \cref{sec:law}, where we explore the role of judges and lawyers in common law legal systems and their distinct functions.
We then introduce the ideas of the rule of law and substantive justice – the pillars on which the judicial system is built and which connect law and morality.\looseness-1

In \cref{sec:proposals}, we explore practical proposals of legal NLP research.
In our view, these proposals can be broken into two groups; 
(1) \textbf{Replacement} with AI; advocates a complete replacement of legal professionals with technology.
(2) \textbf{Augmentation} of judges or lawyers; suggest only partial automation or supplementation of legal tasks with NLP.
Concluding our background sections, in \cref{sec:nlp_background}, we turn to legal NLP and discuss the impact of the shift from symbolic to sub-symbolic AI on the field.

We then proceed in three stages.
In the first stage, \cref{sec:falacy}, we begin by outlining the risks and benefits of legal NLP.
We suggest that current proposals for implementing legal NLP face three technical challenges (1) lack of contextual and social intuition at the trial stage; (2) inability to make controversial moral and political decisions to develop the law; and, (3) inability to justify the decisions to the public. 
Further, even if these technical challenges are resolved, existing proposals pose unavoidable ethical risks. They have the potential to (1) centralise power; (2) produce a more brittle legal system and (3) undermine the accountability of policymakers.\looseness-1

In the second stage \cref{sec:lawyer}, we turn to the main legal actors, the judge and the lawyer, to evaluate their exposure to the risks and benefits.
We suggest the role of lawyers is a more promising domain for legal NLP by showing how the various concerns affect lawyers differently.
In a nutshell, lawyers are not burdened with the task of being lawmakers. 
Instead, their role has more easily measured metrics: 
whether or not they can convince judges and other lawyers of the quality of their arguments.
We recommend research paths towards resolving the technical limitations of automating legal actors – be they lawyers or judges.


In the third stage \cref{sec:voice}, we turn to the role of the judge in contemporary legal NLP research. 
Many existing and popular legal NLP datasets comprise collections of cases, i.e. the text produced by judges.
We demonstrate how a lack of data about lawyers and their arguments can undermine the modelling task built on top of these datasets. 
Therefore, we recommend legal NLP researchers to focus more on the role of the lawyer in the legal system.
Our paper concludes with \cref{sec:related} by contextualising our work in the wider NLP ethics discussion.\looseness-1

Our proposals do not call for a wholesale reorganisation of the field. 
Judges remain an important subject to study to further our understanding of law and language.
Indeed, the datasets built on publicly available case law have opened up research in legal NLP to contemporary deep learning methods.\footnote{See Caselaw Access Project: \url{https://case.law}}
Further, some datasets already implicitly contain the voice of a lawyer, in the form of legal claims, waiting to be disentangled from that of the judge.
In fact, we hope that the shift to a lawyer could, in some cases, be enabled by re-purposing existing datasets.
For example, \citet{chalkidis-etal-2021-paragraph} and \citet{negativeoutcome} predict both lawyer's claims and judicial outcomes, using the same legal dataset – an early step in the right direction.
We, thus, believe that focusing on the lawyer and their interactions with the judge can supplement existing approaches, and inspire new, more robust legal NLP research.
\looseness-1

\section{Legal Background}\label{sec:law}

Our work primarily focuses on common law countries: this includes England, the US, Australia, and India amongst others.
The other major category of legal systems is practised in civil law countries, such as France, Germany, Japan, and Austria \cite{zweigert1992introduction}.
Both systems share a mixture of important differences and commonalities. 
Overall, we focus on common law countries for two reasons. 
First, these are countries with a rich history and a longstanding philosophical debate surrounding them.
Second, common law systems, in particular England and the US, have been highly active in practical proposals for NLP \cite{Cobbe2020} and many of the datasets discussed in \cref{sec:nlp_background} and \cref{sec:voice} involve common law material.\looseness-1

\subsection{Legal Actors}

Under both systems, there is a strict distinction between the role of judges and lawyers. 
Members of the public go to lawyers to frame their needs in legal terms; common examples are legal documents, such as wills, deeds, and contracts. 
In addition, if the client wants to sue someone or is being sued, the lawyer might be asked to assist.
The client specifies their desired outcome, such as avoiding responsibility for causing an injury. 
Lawyers then translate these demands into legal allegations. 
An allegation is a legal argument which suggests the client's preferred dispute outcome is the most consistent with the law \cite{MariQ2022}.
These allegations are usually sent to the opposing party's lawyers before there is any litigation in the form of a legal brief.
The most common result is for the parties to then settle, preventing the case from going to trial \cite{LawReport}.  
If the parties decide to take it to court, the arguments will be repeated in front of a judge.\looseness-1

The central duty of the lawyer is to their client. 
They are responsible to the public only to the extent they commit no crimes and fulfil their professional obligations \cite{Mari2020}.\footnote{There are also lawyers who are employed by the government. For example, public defendants' salaries are paid by the state. However, their obligation in court is to their clients rather than to their employer. Finally, there are quasi-political legal roles, such as that of Attorney Generals in the US, which blur the distinction between a lawyer and a judge.}
Judges, on the other hand, are not hired by members of the public. 
Instead, they work for the state and have a duty to the public to decide cases correctly and fairly \cite{wacks2015law}.
They are employed in large numbers. 
Currently, there are $3,174$ judicial posts in England that are spread out widely across the country \cite{GovReportJudges}, although this is small compared to the number of lawyers.\looseness-1

The judge's role emerges most clearly when parties litigate and go to trial. 
Generally, trials have two stages. 
The first is fact-finding. 
Both sides present their version of the facts to the court and the judge must decide which version they believe is correct. 
A variety of different sources of evidence are used, such as documents, fingerprints, and witnesses \textit{inter alia} \cite{wacks2015law}. 
In common law countries, the fact-finding role can also be carried out by randomly selected members of the public called the jury  \cite{zweigert1992introduction}.\footnote{The prevalence of the jury differs between common law countries: in England, for example, they are present in many criminal cases but few civil cases (cases involving contracts, torts, conveyances). In the US, on the other hand, civil juries are more common in civil cases.}\looseness-1

Second, after the facts are determined, the judge applies the law to determine the outcome of the case. 
At this stage lawyers on both sides present their arguments for what the law is and how it should fit the facts in a way which will benefit their side. 
The judge then has to make a decision and explain it. 
Crucially, the judge is constrained in their decision.
They are only considering the validity of the alleged violations presented to the judge, not the applicability of any law.
If the judgement is considered incorrect, it can be appealed – it goes to a higher court for review \cite{wacks2015law}.\footnote{Typically, when this occurs only the legal arguments are reconsidered and the facts are assumed to be what was decided in the first trial.}

In common law countries, the judge has a law-making role as well. 
Through the doctrine of legal precedent, when a common law judge decides a case, they create a new legal rule; future cases with similar facts to an already decided case must be adjudicated by judges in the same way \cite{garner2009black}.
While past cases are not binding in the civil law systems, they are still used to decide new cases and can carry a certain weight under the principle of consistent law application.\looseness-1

\subsection{The Rule of Law}\label{sec:ruleoflaw}

The Rule of Law is a fundamental political principle recognised by both common and continental jurisdictions underpinning how the legal system operates.
Many different principles fall under the umbrella of Rule of Law.
We will focus on the following four.\looseness-1

\paragraph{Consistency.} 
It is a basic principle that like cases should be treated alike, and that judicial biases should not interfere with legal decisions \cite{fuller}. 
This is not an absolute principle given the law sometimes needs to develop to meet new societal needs, but it ensures the law is predictable and consistent across cases.
It is especially important for lower-tier judges who are expected to dutifully apply the law set out by the courts above, such as the Court of Appeal and Supreme Court. 

\paragraph{Access to Justice.}
Access to justice is another fundamental principle of the legal system \cite{Diver_2020}. 
Legal subjects must be able to gain access to legal advice and have time in court to enforce their legal rights.
Procedural delay, extremely high costs, and geographically sparse courts are all hindrances to access to justice and undermine the rule of law.

\paragraph{Equality Before the Law.} 
A central principle of the modern liberal state is that no one is above the law, including lawmakers \cite{Dicey1979}. 
This means politicians and judges are subject to the rules they make. 
Equality in this way improves the legitimacy of law-making: legal subjects can be assured that there is one legal regime, equally applied to them and the law-maker.
The principle also acts as a check and feedback mechanism for lawmakers: they feel the sting of unfair and unjust legal rules because they are subject to them. 
Further, they can observe firsthand how the rules operate in practice, gaining useful feedback for creating new laws \cite{Dicey1979}.\looseness-1

\paragraph{Comprehensibility.} 
Finally, it is an essential principle that legal subjects can access and understand the law which governs them \cite{fuller}. 
This means the law must be publicly available, understandable, and not contradictory. 
It also requires satisfactory explanations behind why legal decisions were reached. 
This makes it possible for legal subjects to follow the law, as well as giving them the ability to challenge and criticise legal rules \cite{Hart1961, Raz1979}.\looseness-1

\subsection{Substantive Justice}

The rule of law principles relate to the form rather than the content of the law. 
The latter is a matter of substantive justice, which describes if the content of legal rules is morally good or bad.

Law is connected to morality for three reasons \cite{sep-legal-positivism}.\footnote{There is a debate between legal positivists and natural lawyers on whether law \textit{must} be substantively just to qualify as law \cite{Gardner2001}.}  
First, laws deal with inherently moral issues, such as abortion, homicide, and constitutional conflicts. 
Second, laws are often created to pursue moral ends, such as reducing crime, furthering racial and gender equality, and redistributing wealth in society. 
Third, law is a tool used by the state to coordinate itself, and as such can be used for both good and evil at a larger scale than would otherwise be possible. 

Given these connections between law and morality, it is important to ensure lawmakers are accountable to those they govern.
A morally acceptable legal system must have mechanisms of political accountability for lawmakers \cite{Dicey1979}.  
For example, in democratic countries, legislators are held accountable to the voting populace.\footnote{In the US judges are elected, much like politicians.}

Overall, substantive justice is more difficult to assess than the rule of law virtues.
First, its content is inherently contestable. 
What constitutes a morally acceptable law? 
Who should be responsible for holding lawmakers to account? 
These issues are often socially and politically divisive and constitute matters that reasonable minds can differ on. 
Second, what is morally good changes over time. 
What was previously morally acceptable may become unacceptable and vice-versa. 
A good lawmaker must take these changes into account when making law to meet the evolving needs of society.

\section{Legal Automation from a Policy viewpoint}\label{sec:proposals}

A number of proposals exist on the future role of legal NLP \cite{susskind2008end,Alarie2016,AlarieSing,Casey2016,Casey2021,Goldsworthy2019}.
We taxonomise these proposals into two groups: (1) Those that advocate for the \textbf{replacement} of legal actors with technology, and (2) those that advocate for mere \textbf{augmentation}.
The former, more radical view suggests NLP can be used to replace all legal tasks.
The latter, on the other hand, proposes legal NLP can only \emph{supplement} the work of traditional human lawyers and judges, without completely replacing them.\looseness-1

\subsection{Replacement}\label{sec:total}

The most ambitious proponents of legal NLP suggest the entire legal profession can be completely automated \cite{Cobbe2020}. 
There are broadly two approaches to replacement. 
The first suggests a large amount of data, better models and more computational power will allow the creation of highly effective legal NLP models \cite{AlarieSing}. 
Under this view, the models will eventually be able to give the correct answer to any legal question instantly with the benefit of more information than any human lawyer or judge could ever possibly consider \cite{Goldsworthy2019}.
The decisions of this machine lawyer would not be directly contestable by litigants or the accused, removing the need for courts and lawyers.\looseness-1

Under the softer replacement approach, while all judges and lawyers are replaced, a human remains in the loop as the policymaker \cite{AlarieSing}. 
A small number of human policymakers would set general policy objectives, such as to reduce traffic accidents by 40\%. 
Legal NLP models with access to vast quantities of data then design and implement rules to achieve these policy goals \cite{Casey2016}. 
As with the first approach, these rules would be set automatically by the legal NLP model and there would be few, if any, opportunities for appeal or adjudication.
Without adjudication, the need for lawyers is greatly diminished \cite{Casey2021}. 
Instead of hiring a lawyer and going to court, demands for change would have to be directed towards governmental policymakers. 

\subsection{Augmentation}

More conservative proponents view legal NLP as a tool to augment legal work without entirely removing the need for human experts. 
Legal NLP tools of this variety are already here and used widely by legal professionals, in particular when carrying out due diligence, document review, regulatory compliance, and e-discovery. 
In the UK, for example, 48\% of law firms were using AI in their business in 2018 \cite{Miran}. 
Tech startups cater to this extensive use of legal NLP and argue its use can reduce the number of lawyers needed for a particular job \cite{articleCarl}.
One proposal is to use legal question-answering systems to reduce the cost of legal advice, and thus make it far more available. 
Such technology could have important implications for access to justice, allowing clients to obtain legal advice that they might otherwise not be able to afford \cite{Pasquale2019}.\looseness-1

Another proposal is to employ AI to reduce uncertainty in estimating the outcome of potential litigation.
Firms have already begun to use software to predict the likely outcomes of a lawsuit and could use this to advise a client about their risks and liabilities \cite{LawReportWomble}. 
On the public side, several governments have started testing the augmentation of judicial functions, such as sentencing and lower-tier tribunal decision-making in criminal trials. 
The company Equivant, for example, offers an AI product which predicts the re-offending rates of different criminals with weighted factors, and is now used to make parole and detention decisions \cite{Hildebrandt2019}. 
Controversially, automated judges already sentence people in the US \cite{Kehl2017AlgorithmsIT} and China \cite{stern2020automating}.

Finally, some have proposed that augmentation could be implemented in the hierarchy of the appellate court system \cite{Cohen2023}. 
Under this hierarchical view, the lower-tier judges, such as county court judges in England, could be replaced with automated decision-making software.
Higher-tier judges, however, would be retained and would review decisions from the automated decision-making tribunals under an appeal procedure.
Further, if reviewing judgments is cheaper and faster than producing them, it might be possible to have a legal expert review each decision whilst still retaining many of the benefits of automation. 
Such a hybrid system would have an advantage of speed in the lower instances of the court, but ensure that legal precedent continues to be developed by experienced human judges.

\section{Legal NLP}\label{sec:nlp_background}
Legal NLP can trace its origins all the way to the late 1950's \citep{kort_1957, nagel1963applying, lawlor}.\footnote{The field has gone by different names at different times. 
Juris-informatics, legal informatics, legal artificial intelligence or legal NLP are just a few. 
In this paper we stick with legal NLP for consistency.}
One of the earliest symbolic AI systems, Hypo \cite{hypo},
explicitly encoded the principles of case-based reasoning in terms of analogy and difference, often based on hand-extracted features. 
Many others were inspired by it \cite{aleven1997teaching, rissland1991cabaret, branting1991}.\looseness-1

\citet{Aleven2003} originally introduced the task of predicting an outcome of a case as an evaluation task for these models, which has nowadays become the modern benchmark of legal NLP.
For a long time, Issue Based Prediction \cite[IBP; ][]{Ashley2009}, one of the successors of the Hypo model, held the state of the art for this task. 
However, the reliance of the symbolic systems on humans for both processing the input and encoding the changing rules of law hindered their wider adoption.
This became a problem for the deployment of these models since the law can change constantly with every new court decision.\looseness-1

Recently the popularity of machine learning, combined with a desire for more robust models of law, has rejuvenated interest in developing applications for the legal domain.
This shift from symbolic to sub-symbolic legal AI has implications for contemporary legal NLP research.
Instead of focusing on encoding legal reasoning by hand, the aim is to approximate it from legal data.
Since the architectures powering many SOTA approaches in legal NLP are fairly homogeneous, the choice of data the ML models are trained on becomes very important. \looseness-1

The ML tasks trained on legal text include question answering \citep{monroy09}, legal entity recognition \citep{cardellino-etal-2017-legal}, text summarisation \citep{Hachey}, outcome prediction \citep{xu2020distinguish, zhong-etal-2018-legal, aletras} and majority opinion prediction \cite{Valvoda2018}, legal topic classification \cite{nallapati-manning-2008-legal}, court opinion generation \cite{ye-etal-2018-interpretable}, case citation resolution \cite{shaffer-mayhew-2019-legal}, study of legal precedent \cite{valvoda-etal-2021-precedent} or legal consulting \cite{wang-etal-2019-iflylegal}. 
For a comprehensive overview of legal NLP, see \citet{zhong-etal-2020-nlp}.\looseness-1

Yet another strain of work has emerged with an explicit focus on modelling individual judges by sociodemographic and similar features, instead of legal text.
\citet{katz}, who predicts the outcome of the United State Supreme Court case law, primarily use the meta-data of the court. 
Employing the nearest neighbour algorithm, they achieve 70\% accuracy using features such as date, court position in the court hierarchy, judge names and lower court outcome.\looseness-1


\section{Ethical Dimensions of Legal NLP}\label{sec:falacy}

We believe there are both benefits and risks in the use of NLP in the legal domain. 
The benefits are large enough that refusing to use NLP could present a moral failure akin to not using new medical treatments to treat patients.
On the other hand, the ethical risks of NLP are also significant and need to be weighed up against these potential benefits.

\subsection{Benefits of Legal NLP}

We see three major benefits that legal NLP can bring to the rule of law and the substantive content of the law: (1) Accessibility; (2) Consistency; and (3) Capacity. 
We consider each in turn below.

\paragraph{Accessibility.}
Technology could improve access to justice in the following three ways.
First, automated legal services could be created and delivered more quickly than human advice -- enabling fast resolution of legal disputes. 
Second, the price and cost-unpredictability of legal services could significantly decrease if aspects of legal reasoning are automated.
Third, legal NLP could facilitate geographically wider availability of legal services by no longer being bound by the proximity to the physical location of the providers of legal services. 
The last point has been especially useful during the pandemic, when many courts adopted a hybrid or fully online operation regarding some of their hearings.

\paragraph{Consistency.}
There is a high variance in the quality of human-provided legal services.
In practice, this means that different people have access to different quality of legal advice, and related cases might not be decided in similar ways. 
Legal NLP could help play a role in narrowing the gap between different legal actors.
Ideally, this could bring us closer to the ideal of an equal and consistent legal system.\looseness-1


\paragraph{Capacity.}
NLP could help lawyers, judges, and litigants deal with growing legal complexity. 
Over time, more legal sources are produced, at a greater rate, in more detail, and with greater varieties of legitimacy \cite{complexity}. 
Human lawyers have found this growing mass of material difficult to handle \cite{Hildebrandt(2018)}.
The growing volume of legal datasets suggests the quantity and complexity of legal work will only increase. 
However, legal NLP could also help to reduce this complexity and allow for deeper and wider research than any human alone would be capable of \cite{AlarieSing}.
For instance, the capacity of large language models, such as GPT-4 \cite{openai2023gpt4}, to synthesize a scale of data beyond human comprehension might help to address this problem in the future.\looseness-1

\subsection{Risks of Legal NLP}

In addition to benefits, there are also important ethical risks in using NLP in the legal domain. 
These can be grouped into two basic categories: \textbf{technical challenges} and \textbf{inherent challenges}.
The former relate to technical limitations in current legal NLP technology in achieving what skilled lawyers and judges can do \cite{clayton_boyd_2020}. 
These are, in theory, resolvable with better models and improvements in the capacities of NLP.
The latter are inherent challenges which arise with the use of technology in the legal process. 
They capture the ethical and political challenges that remain even if NLP can perfectly replicate legal tasks.

\subsubsection{Technical Challenges}

We discuss three technical challenges of legal NLP: (1) the need for human intuition at trial; (2) the presence of moral and political debate in the legal process; and (3) reason-giving in law.

\paragraph{The Trial Problem.}\josef{added connection to Bender's octopus}
As noted in \cref{sec:law}, a central part of a trial is the fact-finding stage.
For a machine to find evidence, all that is required at this stage is combing through large quantities of textual data \cite{KQT}.
Current NLP techniques excel at this task.
Some types of evidence, however, pose serious problems. 
In many cases, the legally relevant factors are mental states, for instance intent, foresight, and malice, which are inherently subjective.
The judges also have to assess psychological characteristics of defendants or plaintiffs, such as their overall credibility and honesty, and whether they seem violent, reckless, regretful or violent.
Trying to determine automatically if these mental states are present, is a highly nuanced process, which cannot currently be done well from text alone \cite{NeurolawAI,Neurolaw}.
\looseness-1

To be able to use NLP models in court the models will need to uncover the true meaning of the situation at hand.
The lack of grounded understanding is a well-known issue in the wider NLP discourse.
For example, \citet{bender-koller-2020-climbing} discuss the limitations of NLP models in accessing meaning from text alone in their \emph{octopus test}.

\paragraph{The Moral Dimension of Law.} \josef{integrated Leins in here}
Creating substantively just law requires a deep understanding of morality, politics, and changing social conditions. 
A system that has been trained on past legal material, as all current ML-based NLP models are, is likely to be backwards-looking and will not be able to take an active part in evolving the law to meet the needs of society \cite{Markou2020}. 
Further, in common law systems, the legal system has a fluid aspect, and an important judicial role is re-interpreting existing precedents in light of current social needs \cite{Delacroix_2022}.
For example, the highly controversial case of \textit{Roe v Wade} required the US Supreme Court to combine a mixture of modern moral and political considerations with the technical legal interpretation of the Constitution.
The case was made complex not only by the difficult moral debate over abortion, and the political question of appropriate federal state power, but also legal questions over the correct way to interpret the Constitution in a changing society.  
As \citet{talat-etal-2022-machine} and \citet{fraser-etal-2022-moral} point out, there are problems with modelling morality with NLP methods.
Current NLP models of morality only learn about moral situations from a limited context, restricting their capacity to make complex and important moral decisions.\looseness-1

Related to this issue is the work on ethics and legal NLP by \citet{leins-etal-2020-give}.
In their paper, the authors question the ethics of collecting legal data in the first place. 
Pertinent for our discussion above, are their concerns about updating the datasets when legal decisions get reversed or a case is appealed to a higher court.
The danger is that a ML model will always lack behind the latest developments of the legal doctrine.
Furthermore, in the wider NLP discourse, there is an ongoing discussion about how biases in the datasets used to train NLP models can get exaggerated by the model \cite{bender}.
A legal NLP tool acting upon such biases would have severe repercussions in the legal domain. Especially if such a tool is to aid in upholding moral values.\looseness-1

\paragraph{The Justificatory Role of Law.}  \josef{contestability, wider discourse}
Legal decisions are the product of the debate between two parties in a matter that is often highly contentious.
The resolution of this debate is important and must be \textit{comprehensible} and \textit{justifiable} to the public.
Current legal NLP models struggle to give reasons for their decisions.
This raises two problems.
First, without providing a clear explanation, NLP models will be unable to convince the parties to the dispute that a correct decision has been reached.
Second, a central part of the rule of law is the contestability of any legal decision \cite{Delacroix_2022}.
It is important that everyday citizens are able to participate in, understand, and challenge the decisions which govern them. 
A legal system that cannot explain the outcomes of cases could severely restrict this possibility \cite{Hildebrandt2020}.\looseness-1

The issue of contestability of a machine learning system goes beyond legal applications and it has been discussed in the wider NLP literature \cite{mitra-2021-provocation}, as well as in the context of human-computer interaction \cite{contestability_hci} and social computing \cite{contestability_social}.
There are also general limitations and privacy concerns regarding to the use of LLMs which we discuss separately in \cref{sec:related}.

NLP has made huge advances in the past decades, and it hopefully will continue to do so. As NLP advances, we expect the technical challenges to get gradually resolved or at least ameliorated.

\subsubsection{Inherent Challenges}\label{sec:inherent}

We further foresee three inherent challenges of using technology in the legal domain: (1) the centralisation of political power, (2) increased brittleness and (3) lack of accountability. 
Unlike technical challenges, no technical advance in NLP will resolve the inherent challenges.

\paragraph{Centralised Power.}
The benefits of using NLP in the legal system – whether it is accessibility, consistency or capacity – lie in the scalability of the technology.
The underlying assumption is that with technology, there will be fewer human legal actors involved in any single legal process.
The flip side of these benefits, however, is the risk of centralising power.
Without AI, the expertise required to adjudicate legal cases is possessed by judges and lawyers \cite{Cobbe2020}, and it is held in a distributed manner.
These actors work separately from one another and most of them are not coordinated in one group or location at any time. 
They can therefore act as checks on one another.
Replacing and augmenting legal expertise with technology will increase efficiency, but likely result in a situation where there are fewer individual actors in the system. 
This creates the risk of upsetting the current fine-tuned balance:
fewer legal agents could operate with fewer constraints and checks between one another.\looseness-1

\paragraph{Increased Brittleness.} 
The risk of centralising the legal system also raises a danger of increased fragility.
There are two aspects to this.
First, there are the risks associated with making the legal system more dependent on digital infrastructure.
The legal system could become more prone to technical failures such as electrical outages, and cyber-attacks.
\citet{taleb2012antifragile} has described this kind of social system as fragile: it cannot accommodate unusual or extreme events.\footnote{There is naturally a trade off here. Increased accessibility through technology makes the legal system more robust to failures of physical communications for instance.}
The risk of widespread system failure increases proportionally to the level of legal NLP involvement in the legal system.

Second, a major benefit of successful automation is the speed-up of the entire legal process.
A fast legal system will also result in the fast creation and application of new precedents.
This reduces the margin of error in law-making, as there is less time to review, challenge, and test new law before it is re-applied.  
In areas where predictability is crucial, such as land law and contract law, small errors could be devastating if applied across the legal system. 
Therefore, it is essential that decisions must be reached with a high level of reliability.
Legal NLP, therefore, raises a political risk that unjust laws will be applied more quickly and comprehensively, with fewer checks at the application stage.

\paragraph{Lack of Accountability.}
Legal actors must always be held accountable for their decisions, following from the legal principle of equality before the law (\cref{sec:ruleoflaw}).
Currently, legal decisions are made by identifiable human experts. 
This ensures that specific individuals can easily be identified and challenged for their decisions, both by members of the public and by private clients \cite{Diver_2020}.
A risk of introducing legal NLP in the system is that it makes it more difficult to identify who is responsible for specific decisions.

The problem of accountability has been raised with respect to automation at large and in particular in the context of \textbf{self-driving cars} and \textbf{recidivism prediction tools} \cite{gless2016if, legalrecidivism, Ryan2020}.
We can take a lesson from these contexts to identify what might be the risks facing legal NLP in the future.\looseness-1

The context of \textbf{self-driving cars} illustrates the complexity of attribution when it comes to partial automation. 
The liability for self-driving cars typically depends on the level of automation and user control \cite{Ryan2020, Boeglin2015}.
There are six levels of automation.\footnote{The SAE Taxonomy and Definitions for Terms Related to On-Road Motor Vehicle Automated Driving Systems breaks down into six levels (0-5) and can be found \href{https://www.sae.org/standards/content/j3016_202104/}{here}.}
For our purposes, Levels 1-4 are particularly interesting.
In brief, Level 1 (Driver Assistance) and Level 2 (Partial Automation) describe a vehicle with the ability to support the driver in steering, braking and accelerating.
Level 4 (High Automation) is where the car is capable of fully driving itself in some driving modes (for example on a highway).
Level 3 (Conditional Automation) is in between.
The car is somewhat self-driving, but a human must be ready to intervene at any point.

Manufacturers will usually deny responsibility at the lower levels (Level 1 and Level 2). Conversely, they will take responsibility for crashes occurring at Level 4. 
There is considerable uncertainty at Level 3, with the driver sometimes being found liable for reduced damages, whilst in other cases escapes liability altogether \cite{Ryan2020}.
While developed in the context of self-driving cars, the above taxonomy might be a useful method of assessing where the pitfalls of deploying legal NLP might lie.
Level 3 system legal NLP system would be one where automation of an aspect of legal reasoning takes place, but humans are expected to monitor the system for potential failures.\looseness-1

The latter category, \textbf{recidivism prediction tools}, can shed some light on how difficult it is to challenge legal decisions aided by software. 
In the US, algorithmic risk assessment tools have been employed to aid judges in making predictions about the risk of the defendant re-offending before the trial.\footnote{There are many tools currently used. For example, Correctional Offender Management Profiling for Alternative Sanctions (COMPAS); the older Historical Clinical Risk Management-20 (HCR-20); Violent Risk Assessment Guide-Revised (VRAG-R); and Sexual Violence Risk-20 (SVR-20)}
Since the deployment of these systems, the validity of their use in the court has been challenged several times by applicants alleging an infringement of their constitutional rights.

For example, in the case \textit{State v. Loomis}, the accused claimed that the use of risk assessment tools to calculate his criminal sentence breached his rights.
One of his arguments was that the software, due to its proprietary nature, was unaccountable and precluded him from challenging its scientific validity \cite{compas}.
The court rejected this argument on the grounds that, whilst the algorithm was hidden, the data it used was all publicly available.

Worryingly, the courts presiding over these cases have not found the lack of transparency as an infringement to the right of the parties involved \cite{compas}.
Instead, they are mostly sanguine about hidden algorithms and the use of personal characteristics, such as gender, to calculate sentences. 
Their reasoning relies upon the fact that, much like in the case of self-driving cars, where automation is only at Level 1 or 2, there is an identifiable human (the trial judge) who is primarily responsible for the relevant decisions. 

If these early cases are anything to go by, challenging a machine-made decision on the basis of accountability will be difficult, especially where such decisions are made using proprietary technology.
Further, such challenges are apt to get more complicated if the extent of automation is greater, and resolving who should take responsibility in such a situation will require inherently political choices \cite{Ryan2020}.
Should it be the responsibility of the researchers who have designed the legal NLP system, the software engineers who have implemented it for the provider, or the provider of the service \cite{Morison2020}?\looseness-1

Furthermore, as a matter of equality before the law, it is important for those responsible for the decisions to feel the effect of the laws they create.
System administrators are far less exposed than trial judges to the specific decisions and outcomes their tools produce. 
The questions of accountability in legal NLP, therefore, go beyond technical concerns.\looseness-1

\section{Judge vs. Lawyer}\label{sec:lawyer}

Since lawyers and judges both practice law, using technology to automate aspects of their roles should give the public wider access to more consistent and better-informed legal services, and should overall benefit the accessibility, consistency and capacity of the legal system.
Both the automation of the lawyer and the judge would result in these positive effects. 
We will now discuss the roles with respect to the risks, and whether they are also equally balanced between the two primary legal actors.
\looseness-1

\subsection{Technical Challenges: Judge vs Lawyer}

Let's first consider the technical challenges of automating judges and lawyers.

\paragraph{The Trial Problem.}
The judge and the lawyer have different roles during the trial.
The judge is responsible for assessing witness testimonies, a process where they examine inconsistencies, reactions, and emotions from the oral accounts of witnesses and litigants.
The judge, therefore, must be able to read diverse physical cues and contextual social knowledge, skills that are difficult to replicate through textual learning.
These are qualities that current legal NLP tools lack.
The role of the lawyer, on the other hand, is to ascertain their client's side of the story, assess their opposing counsel's evidence, communicate with clients and processing ambiguous facts.
They also need to assess the changing reactions of a jury to arguments.
All of this requires emotional and social sensitivity. 
Finally, lawyers require the ability to decide how to cross-examine witnesses, and create compelling narratives, tailored for the particular judge or jury composition \cite{brooks1996law}. 
Therefore, although their roles are different, lawyers and judges face similar difficulties when it comes to assessing standard legal evidence.\looseness-1 

\begin{recommendation}[Multi-modality]
Legal NLP needs multi-modal approaches, integrating vision and sound, which are necessary for the full automation of judges and lawyers.
\end{recommendation}

\paragraph{The Moral Dimension of Law.}
The requirement of substantive justice in law means lawyers and judges inevitably need to consider ethical and political factors.
Many cases involve morally charged issues and judges are under a duty to the public to reach morally correct decisions. 
Since judges take moral factors into account, lawyers must also include them in their arguments and indicate the ethical context of their client's situation \cite{Liu2022}.
We believe that the role of a judge as a moral arbiter sets a high burden for the full automation of judges.
The metric for morally acceptable law is highly contestable and controversial.
A successful judgment under this criteria must be persuasive to a wide range of groups, ranging from litigants, politicians, academic commentators, and the public generally.
Given the controversial nature of moral judgments, these groups are likely to diverge, and deciding the correct weight to be accorded to each is also contestable. 
This makes it extremely difficult to determine whether a model is successful in making morally good decisions. 

In contrast, a lawyer only needs to successfully appeal to the decision-making of the judge rather than create an argument they themselves believe to be morally sound.
It is not their primary responsibility to ensure law is enforced according to moral standards accepted broadly by different groups in society. 
Rather, their role is to present their client's case in the strongest, most persuasive form possible to convince the judge to decide in their favour.\josef{changes regarding lawyer and morality}
A good argument, under this standard, is one which generally persuades judges and secures victories in lawsuits for the client. 
Lawyer automation, therefore, has a lower technical burden to overcome when it comes to modelling morality.\looseness-1

\begin{recommendation}[Morality]
We suggest following the recommendation of \citet{talat-etal-2022-machine} of including domain experts in developing the moral sense of NLP models.
\end{recommendation}

\paragraph{The Justificatory Role of Law.} 

Both lawyers and judges must explain their decisions. 
However, their justifications serve very different ends.
Judges explain their decisions because of the rule of law principle of \textit{comprehensibility}.
It is important for the members of the public to understand why a judgement was reached, both for its legitimacy and so they can contest it.
Justifying decisions to the public to this level sets a very high technical burden to overcome.
It requires feedback from a wide range of public actors and interest groups. 
Reasonable minds can differ on whether an explanation is comprehensible. 
Furthermore, it takes a long time to assess the quality of the judge's reasoning and whether it has shaped the law in a positive or negative way.
Lawyers on the other hand give justifications for more practical purposes. 
The main test is whether their arguments can persuade judges and other lawyers. 
This means that there is a more easily measured criterion of success:
If other lawyers and judges accept and understand the arguments, the justifications have succeeded.
If they do not, they need improvement.
The burden of automating a judge is therefore higher than that of automating a lawyer.\looseness-1

\begin{recommendation}[Explainability]
    Legal NLP models need to explain their reasoning in a way that allows members of the public to exercise moral and political scrutiny over it.
\end{recommendation}
There are outstanding technical challenges for the full automation of the judge or lawyer. 
In light of these challenges, we believe that augmentation is a more realistic pursuit given the current limitations of legal NLP. 
At the moment, we should leave aside the witness assessment, moral decision making and justifications to human experts. 
Nonetheless, we see the pursuit of the full or partial automation of lawyers as less problematic than that of judges.

\subsection{Inherent Challenges: Judge vs Lawyer}

Now we turn to the challenges that cannot be resolved by technology alone.

\paragraph{Centralised Power.}\josef{natural monopoly vs. mane made one}
The risk of centralising power is much greater if judges are automated than if lawyers are automated.
This is for two reasons: the political power judges possess, and their organisational structure. 
First, as noted in \cref{sec:law}, judges in common law countries possess law-making power, which gives them influence over the lives of everyday people. 
The role of a lawyer is different.
Lawyers do not create new laws, they only make arguments that the judges take into account. 
This considerably limits their ability to change the law. 
Second, judges are more likely to be centralised than lawyers.
As state employees, they are by default employed by a single organisation: the government. 
Additionally, the role of the judge involves high consistency which has tempted some to believe that a single automated decision-making system is desirable in the first place, see \cref{sec:proposals}.

Conversely, lawyers are organised in private firms which compete with one another for the business of clients.
Law firms are free to structure themselves and tailor the advice they give to their clients.
As a result, law firms are more likely to operate independently, choosing their own tools and approaches, whether they are aided by technology or not.
Consequently, it is unlikely that the automation or augmentation of the lawyer profession would be achieved using a single NLP model or tool, and therefore the risk of centralising power is lesser when it comes to their role in the system.\looseness-1

Nonetheless, they are still at risk of centralisation. 
Legal NLP, like other technology sectors, might demonstrate network effects and lock-in, leading to one or two providers dominating the market.
This risk, connected with the potential effect on the development of the law and the accessibility of legal advice, means centralisation remains a concern when automating either actor.\looseness-1

\paragraph{Increased Brittleness.}
Increased centralisation also brings the risk of increased brittleness.
Since judges are more at risk of the former, they are also more susceptible to the latter. 
The importance of role of the judge as a lawmaker magnifies the potential harm that could arise from a failure or malfunction of the tools they use and the aspects of their role which are automated. 
While a human judge can be biased, incompetent, or subversive, they can be checked by other judges.
Deferring aspects of judicial reasoning to a machine threatens self-policing among judges.
Furthermore, the speed and scale at which automated judicial reasoning could cause harm is much greater than in the case of a biased human.

Partial replacement helps mediate some of these issues by introducing human review. 
The extent this mediates the risk will depend on the extent humans are placed in the loop, producing a trade-off between speed, cost, and fragility.
In contrast, lawyers have less influence on the system as a whole, and there is a lower risk of their centralisation. 
Therefore, there is a lower risk of brittleness in the case of their replacement or augmentation.
Clients who are able to choose from a variety of legal service providers can avoid bad actors; faults in tools used by one law firm is therefore unlikely to spill over into the others.\looseness-1 

\begin{recommendation}[Diversity]
    It is necessary to maintain the diversity of adjudicators in order to avoid the risks of power concentration and brittleness. Whether machine or human, having the same automated system for every case poses a danger of systemic malfunction and fragility. 
\end{recommendation}

\paragraph{Lack of Accountability.}
The judge and lawyer are both accountable, but they are accountable to different groups.
The judge is accountable to the public and state officials, whereas lawyers are accountable directly to their clients.
This distinction has implications for the automation of lawyers and judges.
Law firms must meet the needs of their clients; they are liable for malpractice and must deliver services which are competitive in the legal services market. 
As a result, they can naturally act as the site of responsibility should their use of legal NLP fail.
In contrast, judges are accountable to the public and state officials.
There is no law firm to shift the responsibility to, nor a market of legal customers who can evaluate the success of legal NLP in meeting their needs. 

In the most severe case of total replacement, it might become difficult to identify individuals responsible for the decision made by a machine judge.
There are also issues of equality before the law. 
A machine judge which is used to automatically decide cases and, therefore, make new precedents, will have no experience of the law it enacts.
This raises issues of the legitimacy of their decisions.
A machine will lack the basic feedback process present when a judge experiences the effects of the cases they decide on.
Since lawyers do not create law, their legal NLP replacement does not pose this risk.\looseness-1

Even if the use of AI in the court is limited, accountability remains an issue.
If current recidivism prediction software is anything to go by (see \cref{sec:inherent}), it will be difficult to challenge any negative impact a statistically driven tool might have on the judge's decision.
Currently, a claimant has no right to examine the tool that is used to help to decide their legal faith and can only hope that the judge's understanding of the technology is sophisticated enough to account for the biases and errors such a tool might introduce.\looseness-1


\begin{recommendation}[Accountability]
   It is necessary to decide who is accountable for the actions made using legal NLP.
\end{recommendation}

\noindent Resolving the inherent challenges will require careful consideration of how exactly is the technology to be introduced in the courtroom.
While the hybrid augmentation approach seems the most sensible, a danger remains that the human in the loop will overestimate the abilities of the NLP tools and defer to them. We find this particularly concerning when it comes to the automation of the role of a judge.\looseness-1

\section{The Voice of a Judge in NLP}\label{sec:voice}

In the above discussion, we argued that the automation of the lawyer poses fewer challenges than the automation of the judge.
Nonetheless, the tasks lawyers and judges conduct during their work are closely connected.
In this section, we explore several legal NLP tasks and demonstrate how the role of a judge is inadvertently modelled via reliance on judge-created training data.
We then make suggestions on how these tasks could be redefined to prioritise the voice of a lawyer.\looseness-1

\subsection{Legal NLP Tasks}
The three tasks we consider in this section are: Legal Outcome Prediction, Similar Case matching, and Legal Question Answering.
Each task relies on judge-generated text for training data.
Since the major paradigm in legal NLP is ML, the current operationalisations of these tasks turn them into a series of different approaches for modelling a judge.
It is worth noting from the outset that these tasks and their accompanying datasets are not an exclusive list of legal NLP research directions; see \cref{sec:nlp_background}.
Nonetheless, these are tasks used in well-established legal benchmarks, such as LexGLUE and COLIEE \cite{chalkidis-etal-2022-lexglue, coliee}, and thus deserve our attention.

The first task we consider is that of \textbf{Legal Outcome Prediction}.
Given the facts of a case, the task is to predict the outcome of a case \cite{chalkidis-etal-2019-neural}.
While both judges and lawyers are interested in estimating the potential of a case to succeed, they approach this problem from different angles.
As noted above, lawyers are interested in maximising the chance of winning the case through the arguments they create for their clients.
Judges, on the other hand, are interested in establishing a sound precedent and serving justice.
They base their decisions on the situation of the claimant, the arguments presented by the lawyers, and their understanding of the law.
The legal NLP models trained for this task do not correspond to either role above.
Instead, both inputs and outputs the models are trained on are extracted from cases -- a transcript of the judge's reasoning towards the outcome of a case.\looseness-1

To better understand why is this a problem, let's first have a closer look at the model inputs; the facts.
When judges describe the facts of a case, they have already decided on the case outcome.
Therefore, the facts used as input to the models are part of the judge's argument.
A real judge has access to all the evidence presented by the parties to the court.
Legal NLP datasets do not currently contain this information.
Therefore, Legal Outcome Prediction models decide on the outcome given only a small subset of highly curated data when compared to a judge.
Because the fact selection process introduces clues about the outcome of a case, it is easier to predict the outcome from this data than if one had an unbiased full description of what had happened.
This renders the ML operalisation of the task artificially easy.

Furthermore, from a legal standpoint, relying on facts alone is insufficient because the case decision is never made solely on facts.
By relying only on the facts as an input to the model, the Outcome Prediction Task is an artificial task, one that would make sense only if we replaced both legal actors with AI.
We have argued in the previous section why this is undesirable.

Now we can turn to the model outputs; the outcomes.
The most popular legal NLP treatment is to cast the task as a binary decision over all possible laws.
The idea is to predict which laws have been violated, given the facts of a case.
Formally, the model is trained to predict a binary vector $\{1,0\}^K$, where $1$ represents a violation of one of the $K$ laws under consideration.
But it is not the case that a judge passes a decision over every law there is.
Instead, a judge makes a decision with respect to a subset of laws -- those the lawyer has alleged as violated.
We will refer to these as claims from now on.
If claims are not taken into account, $0$ is left ambiguous as it can represent either a law that has been claimed as violated but the judge has decided it is not, or a law that is completely unrelated to the facts at hand.
As \citet{negativeoutcome} point out, the standard operalisation of the task is therefore artificially easy.\looseness-1


The second task is \textbf{Similar Case Matching}.
Under the COLIEE competition Task 1 \cite{coliee}, given a text of a case with redacted references to previous case law, i.e. the precedent, the task is to correctly predict the redacted references (but not their location in the original text), see \cref{tab:examples}.\footnote{Not all datasets for similar case matching have been sourced this way. Some, like \citet{xiao2019cail2019scm}, have been manually created and do not fall in the scope of our critique.}
While both lawyers and judges search for relevant precedents, the data used to train the models to find related cases comes again solely from the judge.
However, judges do not select these cases under some objective metric of relatedness.
Instead, a judge cites cases that support their argument towards the outcome of the case.\looseness-1

This raises two issues.
First, one can safely presume that much like the facts above, the judge-cited cases are an incomplete set of relevant case law.
They almost certainly contain only a portion of the citations that were used by the two parties to the case.
In particular, they are likely to lack some of the precedent that was relied upon by the party that has lost the case.
After all, if the judge agreed with the losing party's argument, she would not decide against it.
This means that by relying on the precedent selected by a judge, the task favours the view of a judge.
However, the precedents that the lawyers have relied upon are equally an indicator of relatedness between any two cases.
These precedents are currently not captured by the COLIEE dataset.\looseness-1

Second, the precedent prediction task is inherently connected to the outcome prediction task above.
If a lawyer wants to claim their client is innocent, they will be looking at a very different legal argument than if they were to claim that their client is guilty.
By ignoring what role in the argument the precedent played, a case retrieved by a Similar Case Prediction model can either support the desired outcome or be against it.
From the perspective of a lawyer, this distinction is crucial when searching for case law to build their argument around.\looseness-1

Finally, we turn to the task of \textbf{Legal Question Answering}.
At first blush, Legal Question Answering might seem like an emulation of a lawyer.
After all, lawyers are paid to answer legal questions.
However, a closer inspection reveals that the questions in the Legal-Domain Question Answering Dataset for example \cite{DBLP:conf/aaai/ZhongXTZ0S20}, are predominantly about the inference of the crime committed, rather than the explanation of legal concepts.
From the $26,365$ questions in the dataset, $16,604$ are case analysis questions, such as the one in \cref{tab:examples}.
The primary difference from the Legal Outcome Prediction task above is that the facts are stylised and simplified.
Take the Legal Outcome Prediction case from \cref{tab:examples} and compare it to the Legal Question Answering example below it.
The tasks are very similar: in both instances, the goal is to predict an outcome of a case, given the facts.
This is, again, an automation of the adjudicatory role of a judge.\looseness-1


\begin{table}[t]
  \centering
  \begin{tabular}{p{7cm}}
    \toprule
    \textbf{Legal Outcome Prediction - \citeauthor{chalkidis-etal-2019-neural}} \\
    \midrule
    \textbf{Facts:} \emph{``Ms Ivana Dvořáčková was born in 1981 with Down Syndrome (trisomy 21) and a damaged heart and lungs. She was in the care of a specialised health institution in Bratislava.
    In 1986 she was examined in the Centre of Paediatric Cardiology in Prague-Motol where it was established that...''} 
    for more see: \href{http://hudoc.echr.coe.int/eng?i=001-93768}{Case of Dvoracek and Dvorackova v. Slovakia}\\
    \midrule
    \textbf{Outcome:} Articles: 2, 6 \\
    \bottomrule\\
    \toprule
    \textbf{Similar Case Matching - \citeauthor{coliee}}\\
    \midrule
    \textbf{Query:} \emph{``The Plaintiff stated that, on the evening of the incident, he was in the telephone area waiting to use the phone. The assailant jumped the queue in an attempt to use the phone.
    The Plaintiff and the assailant "bumped shoulders"..."}\\
    \midrule
    \textbf{Precedent:} 010, 151 \\
    \bottomrule\\
    \toprule
    \textbf{Legal Question Answering - \citeauthor{DBLP:conf/aaai/ZhongXTZ0S20}} \\
    \midrule
    \textbf{Question:} \emph{``What crimes did Alice and Bob commit if they transported more than 1.5 million yuan of counterfeit currency from abroad to China?''}\\
    \midrule
    \textbf{Answer:} Smuggling counterfeit money. \\
    \bottomrule
  \end{tabular}
  \caption{Examples for three popular legal NLP tasks.}
  \label{tab:examples}
\end{table}

\subsection{Proposed Solutions}
We propose two solutions to the above problems: utilising the information about legal claims and collecting new datasets of legal briefs.
Both claims and briefs are a product of a lawyer. 
Therefore, reconstructing the above tasks around datasets of lawyer-generated text turns them from tasks that model a judge into tasks that model a lawyer.\looseness-1

Consider how this shift would be achieved by utilising legal claims.
For the Legal Outcome Prediction task and the related Legal Question Answering task, having the knowledge of both legal claims and outcomes can fix the ambiguity inherent in the binary classification setting.
One could then perform a three-way classification.
The disambiguation is done by subdividing the $0$ category into negative outcomes and null outcomes.
Negative outcomes are the claims that have not succeeded in the court, null outcomes are unclaimed laws.
In their work, \citet{negativeoutcome} have implemented such three-way classification and demonstrated significant improvements over previous approaches.\looseness-1

Alternatively, one could also train the model to predict the claims directly: claim prediction can be reasonably defined as a binary classification task \cite{chalkidis-etal-2021-paragraph}.
This is a simple solution since claim prediction is already the focus of LexGLUE ECtHR Task B \cite{chalkidis-etal-2022-lexglue}.
Because claims are the product of a lawyer, the first approach incorporates the information about a lawyer in the model, while the latter approach models the role of a lawyer.


Let us now consider how collecting legal briefs could help.
Legal briefs are the arguments the lawyers present to the judge on behalf of their client.
They also contain evidence the argument is based on.
A dataset of these briefs could address the limitations of facts as the sole inputs to the Legal Outcome Prediction and Legal Question Answering models.
Specifically, having access to briefs would allow for training models conditioned on the full facts of the case contained in the legal briefs.
Better yet, the model outcome could be conditioned on both the factual description of the situation at hand and the lawyers arguments.
By including the lawyers' arguments as input to the model, the outcome prediction task would stop implicitly assuming a fully automatic legal process, where a judge operates without interaction with a lawyer, but rather a process with two distinct legal actors; lawyers who are developing arguments and judges who are evaluating these arguments.\looseness-1

From a practical perspective, a lawyer could use such a model to estimate their chances of winning a case, given the arguments they develop.
As for the Similar Case Matching task, legal briefs would allow for training models that consider the full spectrum of precedent relevant to a case.
Better yet, with access to this data, one could begin to develop legal NLP models that have the ability to find the precedent relevant to a desired outcome of a case.\looseness-1

In conclusion, the tasks described in this section are restrained by the lack of access to the data produced by a lawyer, which makes them normative to the view of a judge.
The proposed solution is not difficult in practice.
Legal claims are already extracted for the ECtHR dataset, so the shift is a matter of choice of the task we should prioritise when training our models.
While currently no legal briefs dataset exists, the data to build such a dataset is publicly available and we are working on collecting and releasing it to the NLP community.

\section{Related Work}\label{sec:related}

There are three areas of wider NLP research related to our work. 
The first concentrates on the privacy of the data in NLP. 
The second discusses the limitations of large language models (LLMs). 
The third raises concerns about the ethics of legal NLP.
Our work falls in this third category of research.

There is a growing field of research on NLP data privacy \cite{klymenko-etal-2022-differential}. 
Concerns over privacy naturally arise from training NLP models on sensitive data, such as medical records \cite{yadav-etal-2016-deep}. 
These concerns are in particular relevant to legal NLP, since the content of legal documents is often confidential and might additionally contain medical details.
Since training data can be reconstructed from large language models \cite{carlini}, which underpin much of the current research in NLP and legal NLP, data privacy is a pressing issue.
One approach for privacy-preserving techniques for legal NLP is the differential privacy approach, a cryptography-based approach using transformer language models developed by \citet{yin-habernal-2022-privacy}.\looseness-1

Going beyond privacy concerns, machine learning approaches to NLP open the systems up to ethical questions by the use of human-generated data \cite{hovy-spruit-2016-social}. 
Data from the internet, and from social media in particular, may contain biases which, if left unchecked, could adversely affect users on NLP systems trained on such polluted data. 
These concerns extend to legal NLP in two ways. 
First, LLMs harbour biases that can affect legal NLP models built on top of them.
Second, the legal data itself contains biases which we would not want to be replicated in our models \cite{judicial_bias}.\looseness-1

Some researchers have conducted studies on uncovering and quantifying different types of bias \cite{buolamwini2018gender, blodgett-etal-2020-language}. 
Others have started to develop tools to de-bias the models \cite{ravfogel-etal-2022-adversarial, debias}.
With the scale of datasets growing over time, these issues are likely to grow \cite{bender}. 
Furthermore, the ability of current LLMs to consider the languages' social factors is questionable \cite{hovy-yang-2021-importance}.

Finally, while the law can be viewed as a codification of moral reasoning to some extent, very little has been written about the ethics of automating the legal system.
There has been an ongoing critique of NLP research aimed at automating legal sentencing and even calls not to publish this research \cite{leins-etal-2020-give}. 
In response to this criticism, others have warned against the threat of moralism against academic freedoms \cite{tsarapatsanis-aletras-2021-ethical}.
In contrast to the previous work, we focus on the roles of the legal actors and the ethical implications of their automation.\looseness-1

\vspace{-2pt}
\section{Conclusion}
\vspace{-1pt}
In this work, we have evaluated the ethical and practical feasibility of different proposals for NLP research in law.
By surveying a number of popular legal NLP tasks, we identify that the datasets they are built on favour the voice of the judge.
However, because the criteria for what makes a good judge includes moral discretion and political accountability we believe that practical applications of these kinds of NLP models face difficult challenges.
For a machine judge to be successful, it would need to have capabilities far exceeding what is currently available. 
These capabilities include moral and social intuition; the sophisticated ability to give explanations; and, in turn, the ability to receive feedback from members of the public.
Even if these technical challenges were met there are further inherent risks to using legal NLP in the real world.
We have discussed three of them: the centralisation of power, brittleness and lack of accountability.
Comparing the judge and a lawyer, we find the role of a lawyer less susceptible to these challenges.
Our appeal is, therefore, to focus on the voice of the lawyer in the legal NLP datasets.
Furthermore, we believe that overlooking the role of the lawyer hinders current neural approaches for modelling law.
For the academic pursuit of furthering legal NLP research, too much focus on the text produced by judges hides the fruitful and interesting interplay between the lawyers that fuels the legal discourse.

\vspace{-3pt}
\section*{Acknowledgements}
Research for this publication was conducted as part of the NordASIL project financed by Nordforsk grant no. 105178 and the Danish National Research Foundation’s grant no. DNRF169 – Centre of Excellence for Global Mobility Law.

\bibliography{anthology, custom}
\bibliographystyle{acl_natbib}




\end{document}